\documentclass[arxiv]{latex/melba}

\usepackage{mwe} 

\usepackage{amsmath,amsfonts}

\melbaid{YYYY:NNN}  
\doi{10.59275/j.melba.2024-AAAA}
\melbaauthors{Michael W. Rutherford and Fred Prior}  
\email{mwrutherford@uams.edu}
\volume{2}
\firstpageno{1}  
\melbayear{2025}  
\datesubmitted{2025-04-30}  
\datepublished{yyyy-m2-d2}  

\melbaspecialissue{MIDI-B}
\melbaspecialissueeditors{Keyvan Farahani, Fred Prior}

\ShortHeadings{Medical Image De-Identification Resources: Synthetic DICOM Data and Tools for Validation}{Michael W. Rutherford and Fred Prior}

\title{Medical Image De-Identification Resources: \\ Synthetic DICOM Data and Tools for Validation}

\author{
    \firstname Michael W. \surname Rutherford\aff{1}\orcid{0000-0003-2665-753X},
    \firstname Tracy \surname Nolan\aff{1}\orcid{0000-0002-7023-7586},
    \firstname Linmin \surname Pei\aff{2}\orcid{0000-0001-6135-9429},
    \firstname Ulrike \surname Wagner\aff{2}\orcid{0000-0002-3230-5058},
    \firstname Qinyan \surname Pan\aff{3}\orcid{0009-0006-8294-6167},
    \firstname Phillip \surname Farmer\aff{1}\orcid{0000-0003-1448-1346},
    \firstname Kirk \surname Smith\aff{1}\orcid{0000-0002-8735-7576},
    \firstname Benjamin \surname Kopchick\aff{4}\orcid{0000-0003-1125-0155},
    \firstname Laura \surname Opsahl-Ong\aff{4}\orcid{0009-0009-1614-1709},
    \firstname Granger \surname Sutton\aff{5}\orcid{0000-0001-7498-8048},
    \firstname David \surname Clunie\aff{6}\orcid{0000-0002-2406-1145},
    \firstname Keyvan \surname Farahani\aff{7}\orcid{0000-0003-2111-1896},
    \firstname Fred \surname Prior\aff{1}\orcid{0000-0002-6314-5683},
}
\affiliations{
    \num 1 \addr University of Arkansas for Medical Sciences, Little Rock, Arkansas, USA \\
    \num 2 \addr Frederick National Laboratory for Cancer Research, Frederick, Maryland, USA \\
    \num 3 \addr Ellumen, Inc., Silver Spring, MD, USA \\
    \num 4 \addr Deloitte Consulting LLP, New York, NY, USA \\
    \num 5 \addr National Cancer Institute, National Institute of Health (NIH), Bethesda, MD, USA \\
    \num 6 \addr Pixelmed Publishing, Bangor, PA, USA \\
    \num 7 \addr National Heart, Lung, and Blood Institute, National Institute of Health (NIH), Bethesda, MD, USA \\
}

\abstract{
    Medical imaging research increasingly depends on large-scale data sharing to promote research reproducibility and to train Artificial Intelligence (AI) models. Ensuring patient privacy remains a significant challenge for open-access data sharing. Digital Imaging and Communications in Medicine (DICOM), the global standard data format for medical imaging, encodes both essential clinical metadata and extensive protected health information (PHI) and personally identifiable information (PII). Effective de-identification must remove identifiers, preserve scientific utility, and maintain DICOM validity. While tools to perform these tasks are increasingly available, tools for assessing de-identification effectiveness are rare and often rely on subjective reviews, which limits reproducibility and regulatory confidence. To help address this gap, we developed an openly accessible DICOM dataset infused with synthetic PHI/PII and an evaluation framework for benchmarking medical image de-identification workflows. The Medical Image de-identification (MIDI) dataset was built using publicly available de-identified data from The Cancer Imaging Archive (TCIA). It includes 538 subjects (Validation: 216 patients, Test: 322 patients), 605 studies, 708 series, and 53,581 DICOM image instances. These span multiple vendors, imaging modalities, and cancer types. Synthetic PHI and PII were embedded into structured data elements, plain text data elements, and pixel data to simulate real-world identity leaks encountered by TCIA curation teams. Accompanying evaluation tools include a Python script, answer keys (known truth), and mapping files that enable automated comparison of curated data against expected transformations. The framework is aligned with the HIPAA Privacy Rule “Safe Harbor” method, DICOM PS3.15 Confidentiality Profiles, and TCIA best practices. It supports objective, standards-driven evaluation of de-identification workflows, promoting safer and more consistent medical image sharing.
}
\usepackage{array}


\begin{document}

\twocolumn[\maketitle]

\section{Background}

	\enluminure{T}{he} increasing interest in Artificial Intelligence (AI)-driven applications in medical imaging has led to a growing demand for large, diverse, and accessible datasets. However, strict regulations, such as the Health Insurance Portability and Accountability Act (HIPAA) Privacy Rule in the United States and the General Data Protection Regulation (GDPR) in the European Union, require the protection of patient privacy when sharing health information. These legal frameworks significantly constrain the open sharing of clinical imaging data while ensuring patient privacy \citep{de_kok_guide_2023,kondylakis_documenting_2024,phillips_discombobulation_2016,us_dept_of_health_and_human_services_guidance_2012}. De-identification remains a critical challenge, as Protected Health Information (PHI) and Personally Identifiable Information (PII) may be embedded in both metadata and pixel data \citep{diaz_data_2021, langlois_open_2024, moore_-identification_2015}.
    
    In medical imaging, direct identifiers, such as patient names, medical record numbers, and dates of birth, may appear explicitly in metadata fields or burned into pixel data. Indirect identifiers, such as dates of service, institution names, device identifiers, or small geographic locations, may also increase re-identification risk when combined \citep{moore_-identification_2015, us_dept_of_health_and_human_services_guidance_2012, clunie_report_2025}. De-identification efforts must therefore address both types of identifiers across structured metadata and unstructured or visual elements in Digital Imaging and Communications in Medicine (DICOM) \citep{nema_digital_nodate} files.
    
    Manual inspection remains a critical component of de-identification workflows, but is unreliable at scale due to the sheer volume and complexity of clinical imaging data \citep{langlois_open_2024, moore_-identification_2015}. At the same time, unsupervised automated tools often fail to remove all identifiers, especially when dealing with burned-in annotations and unstructured (free text) metadata. As such, effective de-identification requires a hybrid approach combining algorithmic processing with expert human oversight \citep{clark_cancer_2013, clunie_report_2025, langlois_open_2024}.
    
    Despite the widespread need, evaluation of de-identifi\-cation success is not standardized. While some large-scale resources, such as The Cancer Imaging Archive (TCIA), employ human expert-driven validation pipelines \citep{bennett_reengineering_2018, clark_cancer_2013}, most research groups lack tools and resources for verifying that identifying information has been completely removed. Informal methods, such as spot-checking or relying on tool defaults, risk exposing identifying information or alternatively degrading scientific utility, potentially undermining reproducibility \citep{kondylakis_documenting_2024, phillips_discombobulation_2016}. As large data sharing becomes increasingly important, there is a growing need for automated and reproducible tools to objectively evaluate de-identification performance \citep{kushida_strategies_2012, phillips_discombobulation_2016}.

\section{Summary}

    Developed under the National Cancer Institute's (NCI) Medical Image De-Identification (MIDI) initiative aimed at automating the de-identification process, this resource introduces a synthetic dataset, hereafter referred to as the MIDI dataset, and an accompanying validation tool designed to evaluate the effectiveness of medical image de-identification workflows. It supports reproducible, attribute-level scoring of PHI/PII removal from structured and unstructured metadata and pixel data, enabling users to objectively assess whether de-identification processes have been successful.
    
    The vision of the MIDI Dataset as a resource is to promote measurable, reproducible, and standards-aligned assessment of PHI/PII removal in clinical imaging. The mission is to enable researchers, developers, and data custodians to benchmark and improve de-identification methods by providing an evaluation framework grounded in synthetic data and answer-key-driven validation built from regulations, standards, expert-level experience, and best practices in preserving the research value of imaging data.
    
    The scope of this resource is broad. Synthetic PHI/PII was inserted into both DICOM structured and unstructured data elements, as well as pixel data of images representing multiple vendors and image acquisition modalities. This enables assessment of the detection and redaction of identifiers by comparison against a known ground truth, in alignment with HIPAA Privacy Rule “Safe Harbor” \citep{us_dept_of_health_and_human_services_guidance_2012} and DICOM PS3.15 Confidentiality Profiles \citep{nema_digital_nodate-1}.
    
    Synthetic data offers an inherently privacy-safe strategy for evaluating de-identification pipelines, allowing rigorous evaluation without exposing patient or personal information \citep{dankar_fake_2021,dumont_schutte_overcoming_2021,pezoulas_synthetic_2024}. Although synthetic data is increasingly used in medical AI research, few resources are tailored specifically to evaluate de-identification performance \citep{gonzales_synthetic_2023, pezoulas_synthetic_2024}. To address this gap, we developed a structured synthetic dataset with embedded PHI/PII, an answer key, and a validation script. This resource complements our earlier work, \cite{rutherford_dicom_2021}, where we presented a smaller synthetic dataset and aligns with priorities identified by the NCI Medical Image De-Identification (MIDI) Initiative Task Group \citep{clunie_report_2025}.
    
    The intended audience for this resource is developers and customers of de-identification tools who work with sensitive imaging data. The resource is designed to be usable by individuals with basic familiarity with Python and command-line tools. It requires a standard Python environment and depends on commonly used scientific libraries. The tool is packaged with documentation and can be executed from the command line with minimal setup. Its design supports development, reproducibility, scalability, and comparability across different de-identification tools and methods.

\section{Discussion}

    By releasing the synthetic MIDI dataset and validation script, we aim to provide the research community with practical tools to test, refine, and benchmark de-identification workflows. This framework enables regulatory compliance with HIPAA Privacy Rule “Safe Harbor” and DICOM standards. It also helps identify areas that may require manual review by identifying edge-case transformations such as partially redacted tags or inconsistent Unique Identifier (UID) replacements. In addition, the framework addresses the need to handle private data elements and burned-in pixel PHI/PII using automated redaction methods. Although the primary focus is on enabling secure and compliant data sharing, and not curation per se, the methods described may also be useful in supporting broader goals related to reproducibility and data quality assurance by detecting missing tags, UID mismatches, or formatting errors that may affect downstream use.
    
    Structured direct and indirect identifier insertion and answer-key-based evaluation provide a consistent and repeatable mechanism for assessing de-identification performance across different tools and institutions. This approach supports reproducible data curation practices. It can serve as a practical testbed for iterative development, regression testing, and evaluation of de-identification workflows. Some of the early uses of the MIDI dataset and evaluation tools included the development of the MIDI pipeline for automated image de-identification, and the MIDI Benchmark (MIDI-B) Challenge (held at MICCAI 2024), which invited tool developers to benchmark their de-identification tools. These activities are described in accompanying papers in this MELBA special issue on MIDI. This resource builds upon best practices emphasized in recent literature. The MIDI Task Group underscored the importance of transparency, documentation, and reproducibility in de-identification workflows \citep{clunie_report_2025}. Others have noted that informal or manual validation remains the norm in many institutions \citep{kondylakis_documenting_2024}, with limited tools available to support benchmarking against known ground truth. Our tool addresses these gaps.

    \subsection{Limitations}    
    
        Our approach relies primarily on hand-crafted rule-based logic for both synthetic data generation and evaluation. Rule-based strategies offer transparency and reproducibility because all modifications are explicitly defined, interpretable, and consistently applied across datasets. This design prioritizes auditability over realism. However, rule-based approaches may not capture the full complexity of real-world imaging environments. Although we have attempted to be exhaustive based on TCIA experience, certain advanced modalities, vendor-specific sequences, and custom private elements may be underrepresented in the current dataset. As technologies and standards continue to evolve, new needs will continue to emerge that should be accounted for. While generative models trained on clinical data distributions might theoretically enable the creation of more realistic and varied synthetic DICOM headers and pixel data that better reflect specific pathologies or vendor-specific modalities \citep{dumont_schutte_overcoming_2021, pezoulas_synthetic_2024}, our emphasis in this work was on building a transparent and reproducible benchmark. Future work could explore generative methods to address complexities that are difficult to model using rule-based logic alone.
        
        The DICOM standard for de-identification \citep{nema_digital_nodate-1} is exhaustive, but does not cover all cases where PHI/PII may be found, particularly in non-standard or inappropriate uses of DICOM data elements. Many DICOM elements allow free-text input by operators, creating the possibility for identifiers to appear in unexpected places. In practice, rule-based curation practices may vary depending on the curator’s interpretation and the intended use of the dataset. Private data elements, which are manufacturer-specific or institution-specific attributes outside the public DICOM standard, present a particular challenge because they are not consistently defined and may contain PHI/PII depending on how they are used. Handling these cases often relies on institutional knowledge and experience, balancing the need to preserve information for research utility. The ground truth answer key presented in this work was developed based on TCIA curation practices, incorporating years of experience curating public imaging datasets for research use. Actions needed for some data elements are based on consensus value judgments rather than a fixed de-identification rule, particularly for private elements. The specific action can change at the discretion of the curator making the decision. While the use of DICOM options like Retain Safe Private helps preserve benign (so-called “known safe”) private elements, broader support for vendor-specific metadata is needed to maximize research utility without compromising privacy.
        
        Although our “validation script” evaluates the application and consistency of important actions, such as removal of PHI/PII in structured and unstructured text, and modification of dates, it currently verifies retention of DICOM referential integrity only for UIDs and Patient IDs. It does not evaluate whether dates that are shifted remain coherent across related imaging objects, which if not performed correctly, may risk the loss of clinically relevant context. 

        In addition, a small number of discrepancies users may encounter stem from issues present in the original datasets used as the foundation for synthetic data generation. In some cases, data inconsistencies related to DICOM data element presence and element value presence were inherited into the synthetic dataset. While these issues reflect real-world data conditions and were retained to preserve authenticity, they complicate the evaluation of de-identification accuracy and are counted as failures if no action is taken to correct them. Additionally, files may be removed during the curation process to maintain dataset quality and DICOM compliance. Although these removals may be justified for curation purposes, they will introduce discrepancies for all ground truth actions being evaluated except text removal. These cases highlight an important limitation of strict answer-key-based evaluation: it may not fully account for curator judgment or necessary interventions performed to improve dataset quality.        

\section{Resource Availability}

    In this manuscript, we present multiple resources, including two datasets, along with software tools and accompanying files for evaluating the effectiveness of de-identification methods. 

    \subsection{Data/Code Location}   
    
        This MIDI data resource includes (1) a dataset populated with synthetic PHI and PII, (2) an answer key (known truth) in the form of expected changes to each file in the synthetic dataset, (3) a “validation script” that compares synthetic data that has been processed by a system under test against the answer key to detect, record, and quantify differences, (4) a TCIA-curated dataset with PHI and PII redacted from the synthetic dataset for comparison purposes, and, (5) Patient ID and UID mapping files that cross-reference identifiers that changed during the curation of the synthetic dataset to produce the curated dataset. These MIDI data resources are publicly available from TCIA at \url{https://doi.org/10.7937/cf2p-aw56}.
        
        The MIDI evaluation software for determining whether a dataset has been de-identified as intended is available from Github at \sloppy \url{https://github.com/CBIIT/MIDI_validation_script}.

    \subsection{Potential Use Cases}
    
        A developer creating de-identification software requires evaluation to ensure correct functionality. The MIDI synthetic data and de-identification evaluation tools reported here enable the creation and testing of methods used to de-identify medical image data.        
        
        A customer evaluating de-identification software requires benchmarking to confirm functionality. The MIDI synthetic data and de-identification evaluation tools provided allow for such validation and verification to confirm that software meets requirements and specifications and that it fulfills its intended purpose.

    \subsection{Licensing}
    
        The MIDI synthetic dataset with inserted PHI and PII, TCIA curated dataset, answer keys, and mapping files are shared under a Creative Commons international Attribution license 4.0. The source files for the synthetic data were initially provided under one of the following licenses: CC-BY-3.0, CC-BY-4.0, and CC-BY-4.0 (Noncommercial). These licenses cover scans of human subjects and phantoms with no direct traceability to the original subjects. The data in the current work have been transformed sufficiently that the CC-BY-4.0 license, typical of current public-access datasets in TCIA, is appropriate for these resources, without any restrictions on commercial use.  
        
        The MIDI evaluation software uses the Apache v2 License, which is a permissive free software license that permits the use of the software for any purpose, redistribution of the software, modification, and the right to distribute modified versions of the software.  

    \subsection{Ethical Considerations}
    
        Data for inclusion in this project was selected from publicly available collections hosted in TCIA. TCIA collections were submitted over the past 14 years, and all data were collected under consent or an appropriate waiver of consent. TCIA complies with the University of Arkansas for Medical Sciences (UAMS) Institutional Review Board protocol \#205568.

\section{Methods}
    We created the MIDI synthetic dataset and supporting tools to evaluate the effectiveness of medical image de-identification methods. Here we will outline each component.
	
    \subsection{Data Details}

        \begin{table}[htbp]
    \centering
    \caption{MIDI Synthetic Dataset Characterization}
    \begin{tabular}{llrrrr}
    \hline
    \textbf{Dataset} & \textbf{Patients} & \textbf{Studies} & \textbf{Series} & \textbf{Instances} \\
    \hline
    Validation & 216 & 241 & 280 & 23,921 \\
    Test & 322 & 364 & 428 & 29,660 \\
    \hline    
     & \textbf{538} & \textbf{605} & \textbf{708} & \textbf{53,581} \\
    \hline
    \end{tabular}
  \label{tab:synth_data_counts}    
\end{table}

        \begin{table*}[t]
  \centering
  \caption{MIDI Synthetic Data Characterization by Modality}
    \begin{tabular}{lrrrrp{11.5em}p{13em}}
    \hline
    \textbf{Modality} & 
    \textbf{Patients} & 
    \textbf{Studies} & 
    \textbf{Series} & 
    \textbf{Images} & 
    \textbf{Anatomy (\# Studies)} & 
    \textbf{Manufacturer (\# Studies)} \\
    \hline
    CR    & 55    & 61    & 63    & 66    & 
    ABDOMEN (1), CHEST (54), PORT CHEST (6) & 
    BLANK (15), Agfa (36), GE Healthcare (2), KONICA MINOLTA (1), Philips Medical Systems (5), Siemens (2) \\
    \hline
    CT    & 100   & 123   & 126   & 11,839 & 
    BLANK (10), ABDOMEN (8), BLADDER (4), BREAST (30), CHESTABDOMEN (1), CHESTABDPELVIS (1), COLON (26), LUNG (32), OVARY (9), PANCREAS (2) & 
    BLANK (2), ADAC (9), CMS, Inc. (2), ELEKTA / MIM Software (1), GE Medical Systems (46), Philips (5), Philips Medical Systems (4), Siemens (33), Siemens Healthcare (1), Varian Imaging Laboratories, Switzerland (20) \\
    \hline
    DX    & 54    & 59    & 62    & 92    & 
    ABDOMEN (2), CHEST (57) & 
    BLANK (28), GE Healthcare (1), GE Medical Systems (18), Philips (3), Philips Medical Systems (9) \\
    \hline
    MG    & 62    & 62    & 63    & 75    & 
    BREAST (61), Left Breast (1) & 
    BLANK (8), HOLOGIC, Inc. (51), LORAD (3) \\
    \hline
    MR    & 131   & 140   & 148   & 8,918 & 
    BLANK (2), ABDOMEN (1), BLADDER (2), BREAST (104), CERVIX (3), KIDNEY (9), PELVIS (9), PROSTATE (10) & 
    BLANK (5), Confirma Inc. (4), GE Medical Systems (95), Philips Medical Systems (11), Siemens (26) \\
    \hline
    PT    & 73    & 99    & 122   & 32,258 & 
    BLANK (17), BREAST (77), CHEST (3), LUNG (1), THYROID (1) & 
    BLANK (11), CPS (4), GE Medical Systems (77), Philips Medical Systems (6), Siemens (1) \\
    \hline
    SR    & 52    & 56    & 64    & 64    & 
    BLANK (56) & 
    GE Medical Systems (5), PixelMed (20), QIICR (29), Siemens Corporate Research (2) \\
    \hline
    US    & 60    & 61    & 61    & 269   & 
    BLANK (59), PANCREAS (2) & 
    Eigen (59), GE Healthcare (1), Siemens (1) \\
    \hline
    \textbf{Total} & \textbf{587}   & \textbf{661}   & \textbf{709}   & \textbf{53,581} &       &  \\
    \hline
    \end{tabular}
  \label{tab:synth_data_mod_body_man_counts}
\end{table*}   

        \subsubsection{MIDI Synthetic Dataset}
        
            The MIDI synthetic dataset was constructed using hand-crafted rules-based methods to insert synthetic PHI and PII into a diverse set of DICOM images. The dataset includes 538 patients, 605 studies, 708 series, and 53,581 DICOM instances. We selected de-identified images from TCIA \citep{clark_cancer_2013} to represent widely used imaging modalities (including CT, MR, PET, US, CR, DX and MG, as well as some SR) and multiple vendors. Synthetic values simulating direct and indirect identifiers were inserted into both standard and private data elements. These include patient demographics, addresses, and dates. The values simulate identity leakage scenarios encountered and logged during routine TCIA curation activities. For the purposes of a de-identification challenge, the dataset was split into validation and test subsets.

            \textbf{Table \ref{tab:synth_data_counts}} displays the total counts of the MIDI synthetic dataset, listing the number of patients, studies, series, and instances for both the validation and test subsets. \textbf{Table \ref{tab:synth_data_mod_body_man_counts}} breaks down these counts by DICOM modality, displaying study counts for both the anatomy and manufacturers represented in the total MIDI dataset. Patients can have one or more studies, each study can contain one or more series, with each series representing a modality, so patients may be represented in multiple modality columns. This results in the totals, with the exception of instances, not being consistent between \textbf{Table \ref{tab:synth_data_counts}} and \textbf{Table \ref{tab:synth_data_mod_body_man_counts}}. 

        \subsubsection{MIDI Curated Dataset}
            To demonstrate reference de-identification practices, the MIDI synthetic dataset was processed through TCIA’s established curation pipeline, which applies the DICOM Basic Application Confidentiality Profile and additional profile options. The resulting curated dataset contains 538 patients, 605 studies, 708 series, and 53,577 DICOM instances, four fewer instances than the synthetic dataset, reflecting the removal of files due to the removal of files from a series that contained mixed modalities. These files were removed to correct the anomaly and ensure DICOM compliance. \textbf{Table \ref{tab:curated_data_counts}} summarizes the number of patients, studies, series, and instances in the validation and test subsets in the curated dataset. As part of the curation process, HIPAA Privacy Rule “Safe Harbor” identifiers were removed or transformed, and all dates were shifted consistently to preserve temporal relationships. 

            \begin{table}[htbp]
  \centering
  \caption{MIDI Curated Dataset Characterization}
    \begin{tabular}{llrrrr}
    \hline
    \textbf{Dataset} &  \textbf{Patients} &  \textbf{Studies} &  \textbf{Series} &  \textbf{Instances} \\
    \hline
     Validation & 216 & 241 & 280 & 23,921 \\
     Test & 322 & 364 & 428 & 29,656 \\
    \hline     
       &  \textbf{538} & \textbf{605} & \textbf{708} & \textbf{53,577} \\
    \hline
    \end{tabular}%
  \label{tab:curated_data_counts}%
\end{table}%

            TCIA de-identifies DICOM images according to standards defined in PS3.15 of the DICOM standard, using Application Confidentiality Profiles to remove or modify PHI/PII while preserving research utility. TCIA applies the Basic Application Confidentiality Profile along with several profile options to enhance both privacy protection (remove more) and data usefulness (retain more): the Clean Pixel Data Option requires the removal of burned-in PHI/PII in the pixel data; the Clean Descriptors Option requires the removal of PHI/PII from free-text fields; the Retain Longitudinal With Modified Dates Option requires preservation of temporal relationships through consistent date shifting; the Retain Patient Characteristics Option requires retention of traits such as age, sex, height and weight, which though potential indirect identifiers, may be essential for research utility (e.g., computing PET SUV); and the Retain Safe Private Option allows for retention of “known safe” private data elements confirmed to be PHI/PII-free. 
            
            In addition to these standard profiles and options, TCIA conducts internal human reviews of the data, using tools to manually inspect all header values and pixel data to ensure PHI/PII is fully removed or sanitized (cleaned) \cite{bennett_reengineering_2018}.  To ensure scientific value is preserved, TCIA attempts to retain any value of possible significance for unanticipated secondary re-use. This proves difficult with regard to private data elements, since by definition these are less standardized and may be poorly documented by the vendor. Though common practice may be to remove them entirely, instead, TCIA catalogs each private element encountered, along with the action taken (removed, retained, or modified). Over decades TCIA has constructed a consensus-based reference to guide future curation decisions called the \textit{TCIA private tag knowledgebase} \cite{moore_-identification_2015,bennett_reengineering_2018}. This knowledgebase is publicly available at \url{http://wiki.cancerimagingarchive.net/display/Public/Submission+and+De-identification+Overview}.

        \subsubsection{MIDI Answer Key}
            During the process of inserting synthetic PHI/PII into the de-identified dataset selected from TCIA, each modification was logged and associated with a specific action that would reverse the insertion, forming the basis of the answer key. These actions, summarized in \textbf{Table \ref{tab:answer_actions}}, are modeled after TCIA’s curation best practices and define the expected transformations for each DICOM standard and private data element affected.
            
            \begin{table}[htbp]
  \centering
  \caption{MIDI Answer Key Action Types}
    \begin{tabular}{lp{15em}}
    \hline
    \textbf{Action} & \textbf{Description} \\
    \hline
    date\_shifted & The date was changed. \\
    patid\_consistent & The patient id is consistent with the patient id mapping file. \\
    pixels\_hidden & The burned-in pixels within the specified coordinates are hidden. \\
    pixels\_retained & The pixels are unchanged when they should not be changed. \\
    tag\_retained & The tag itself is retained and present in the DICOM image set. \\
    text\_notnull & The value of the tag is not null or zero length value. \\
    text\_removed & The text specified was removed from the tag value. \\
    text\_retained & The text specified was retained in the tag value. \\
    uid\_changed & The UID was changed. \\
    uid\_consistent & The UID is consistent with the UID mapping file. \\
    \hline
    \end{tabular}%
  \label{tab:answer_actions}%
\end{table}%
            
            Many actions, such as date\_shifted, uid\_changed, uid\_consistent, and patid\_consistent, are applied programmatically. It is straightforward to confirm that identifiers have been altered or the actions have been applied consistently across related files. Other actions, such as text\_retained and text\_removed, are more context-dependent. Changes to these will reflect curator judgment, making them non-binary and more challenging to evaluate.

            \begin{table}[htbp]
  \centering
  \caption{MIDI Answer Key Categories}
    \begin{tabular}{ll}
    \hline
    \textbf{Category} & \textbf{Description} \\
    \hline
    HIPAA & Based on the HIPAA privacy rules \\
    DICOM & Based on the DICOM Standard \\
    TCIA  & Based on TCIA’s curation practices \\
    \hline
    \end{tabular}%
  \label{tab:answer_categories}%
\end{table}%
            
            These decisions are captured in the answer key. To support interpretation and downstream analysis, each answer key entry is also categorized by the nature and purpose of its action. These categories, summarized in \textbf{Table \ref{tab:answer_categories}}, are grouped into HIPAA, DICOM, and TCIA.

            Each category is further broken down into subcategories (\textbf{Table \ref{tab:answer_subcategories}}), with bracketed identifiers that link to supporting tables. For HIPAA-related elements, \textbf{Table \ref{tab:hipaa_safe_harbor}} lists Safe Harbor Codes. The DICOM category references both DICOM attribute types (\textbf{Table \ref{tab:dicom_types}}) and de-identification action codes (\textbf{Table \ref{tab:p15_actions}}) from DICOM PS3.15. The TCIA category includes actions based on manual curator review, TCIA’s private tag knowledgebase, manual curator review, and multiple PS3.15 options listed in \textbf{Table \ref{tab:p15_options}}.
            
            \begin{table}[htbp]
  \centering
  \caption{MIDI Answer Key Subcategories}
    \begin{tabular}{lp{11.5em}}
    \hline
    \textbf{Subcategory} & \textbf{Description} \\
    \hline
    HIPAA-\textbf{\{A\}} & HIPAA Privacy Rule Safe Harbor requirements \\
    DICOM-IOD-\textbf{\{B\}} & DICOM Standard IOD \ Requirements \\
    DICOM-P15-\textbf{\{C\}} & DICOM Standard PS3.15 \\
    TCIA-P15-\textbf{\{D\}}-\textbf{\{C\}} & TCIA’s use of DICOM Standard PS3.15 Options \\
    TCIA-PTKB-\textbf{\{C\}} & TCIA’s Private Tag Knowledgebase \\
    TCIA-REV & TCIA Curator Review \\
    \hline
    \end{tabular}%
  \label{tab:answer_subcategories}%
\end{table}%

            \begin{table*}[t]
  \centering
  \caption{HIPAA Privacy Rule Safe Harbor Codes \textbf{\{A\}}}
    \begin{tabular}{lp{40em}}
    \hline
    \textbf{Code} & \textbf{Description} \\
    \hline
    \textbf{A} & The \textit{name(s)} of the individual patient, their family members and employers, household member, and other close connections that could be used to identify the subject. \\
    \textbf{B} & All \textit{geographic subdivisions} smaller than a state, including street address, city, county, precinct, ZIP code, and their equivalent geocodes, except for the initial three digits of the ZIP code, if, according to the current publicly available data from the Bureau of the Census: \\
    \textbf{C} & All elements of \textit{dates} (except year) for dates that are directly related to an individual, including birth date, admission date, discharge date, death date, and all ages over 89 and all elements of dates (including year) indicative of such age, except that such ages and elements may be aggregated into a single category of age 90 or older \\
    \textbf{D} & \textit{Telephone numbers} at which the patient can be reached (home, mobile, work, etc.) \\
    \textbf{E} & \textit{Fax numbers} at which the patient may be sent documents, including home and work \\
    \textbf{F} & \textit{Email addresses} at which the patient can be reached, including personal and work \\
    \textbf{G} & The patient’s \textit{social security number} \\
    \textbf{H} & The patient's \textit{medical record numbers} \\
    \textbf{I} & Health plan \textit{beneficiary numbers} \\
    \textbf{J} & \textit{Account numbers} \\
    \textbf{K} & \textit{Certificate/license numbers} \\
    \textbf{L} & \textit{Vehicle identifiers} and serial numbers, including license plate numbers \\
    \textbf{M} & \textit{Device identifiers} and serial numbers \\
    \textbf{N} & \textit{Web Universal Resource Locators} (URLs) \\
    \textbf{O} & \textit{Internet Protocol (IP) addresses} \\
    \textbf{P} & \textit{Biometric identifiers}, including finger and voice prints \\
    \textbf{Q} & \textit{Full-face photographs} and any comparable images \\
    \textbf{R} & Any \textit{other unique identifying number}, characteristic, or code \\
    \hline
    \end{tabular}
  \label{tab:hipaa_safe_harbor}
\end{table*}
            
            \begin{table}[t]
  \centering
  \caption{DICOM Attribute Types \textbf{\{B\}}}
    \begin{tabular}{lp{18em}}
    \hline
    \textbf{Type} & \textbf{Description} \\
    \hline
    \textbf{1} & Required to be in the SOP Instance and shall have a valid value. \\
    \textbf{2} & Required to be in the SOP Instance but may contain the value of "unknown", or a zero-length value. \\
    \textbf{3} & Optional. May or may not be included and could be zero length. \\
    \textbf{1C} & Conditional. If condition is met, then it is a Type 1 \\
    \textbf{2C} & Conditional. If condition is met, then it is a Type 2 \\
    \hline
    \end{tabular}%
  \label{tab:dicom_types}%
\end{table}%
            
            \begin{table*}[htbp]
  \centering
  \caption{DICOM PS3.15 Action Codes \textbf{\{C\}}}
    \begin{tabular}{lp{40em}}
    \hline
    \textbf{Code} & \textbf{Description} \\
    \hline
    \textbf{D} & replace with a non-zero length value that may be a dummy value and consistent with the VR \\
    \textbf{Z} & replace with a zero-length value, or a non-zero length value that may be a dummy value and consistent with the VR \\
    \textbf{X} & remove \\
    \textbf{K} & keep (unchanged for non-sequence attributes, cleaned for sequences) \\
    \textbf{C} & clean, that is replace with values of similar meaning known not to contain identifying information and consistent with the VR \\
    \textbf{U} & replace with a non-zero length UID that is internally consistent within a set of Instances \\
    \textbf{Z/D} & Z unless D is required to maintain IOD conformance (Type 2 versus Type 1) \\
    \textbf{X/Z} & X unless Z is required to maintain IOD conformance (Type 3 versus Type 2) \\
    \textbf{X/D} & X unless D is required to maintain IOD conformance (Type 3 versus Type 1) \\
    \textbf{X/Z/D} & X unless Z or D is required to maintain IOD conformance (Type 3 versus Type 2 versus Type 1) \\
    \textbf{X/Z/U} & X unless Z or replacement of contained instance UIDs (U) is required to maintain IOD conformance (Type 3 versus Type 2 versus Type 1 sequences containing UID references) \\
    \hline
    \end{tabular}%
  \label{tab:p15_actions}%
\end{table*}%
            
            \begin{table}[htbp]
  \centering
  \caption{DICOM PS3.15 Options \textbf{\{D\}}}
    \begin{tabular}{lp{18em}}
    \hline
    \textbf{Option} & \textbf{Description} \\
    \hline
    \textbf{BASIC} & Basic Profile \\
    \textbf{SAFE} & Retain Safe Private \\
    \textbf{DEV} & Retain Device Identify \\
    \textbf{PAT} & Retain Patient Characteristics \\
    \textbf{MOD} & Retain Longitudinal with Modified Dates \\
    \textbf{DESC} & Clean Descriptors \\
    \textbf{PIX} & Clean Pixel Data \\
    \hline
    \end{tabular}%
  \label{tab:p15_options}%
\end{table}%

            The structure of the answer key is shown in \textbf{Table \ref{tab:answer_key}}. Each record includes the SOP Instance UID, information entity scope (study, series or instance), data element tag path (since data elements may be nested), data element tag name, synthetic file value, expected action, and action-associated text value (part of the file value that the action is to be performed on, which may be only part of the text).
            
            \begin{table*}[t]
  \centering
  \caption{MIDI Answer Key}
    \begin{tabular}{lllllll}
    \hline
    \textbf{UID} & \textbf{Scope} & \textbf{Tag} & \textbf{Name} & \textbf{File Value} & \textbf{Action} & \textbf{Action Text} \\
    \hline
    2.1... &  Instance  &  (0020,0062)  &  Image Laterality  &   &  text\_notnull  &    \\
    2.1... &  Instance  &  (0008,0050)  &  Accession Num  &  20180805E673674  &  text\_removed  &  20180805E673674  \\
    2.1... &  Instance  &  (0008,0090)  &  Referring Phys  &  TURNER\^{}PAUL  &  text\_removed  &  TURNER\^{}PAUL  \\
    2.1... &  Instance  &  (0010,0010)  &  Patient's Name  &  SANCHEZ\^{}TIM  &  text\_removed  &  SANCHEZ\^{}TIM  \\
    2.1... &  Instance  &  (0010,0020)  &  Patient ID  &  1814567196  &  text\_removed  &  1814567196  \\
    2.1... &  Instance  &  (0010,0030)  &  Patient's Birth &  19720701  &  text\_removed  &  19720701  \\
    2.1... &  Instance  &  (0008,0020)  &  Study Date  &  20180805  &  date\_shifted  &  20180805  \\
    2.1... &  Instance  &  (0008,0021)  &  Series Date  &  20180805  &  date\_shifted  &  20180805  \\
    2.1... &  Instance  &  (0008,0022)  &  Acquisition Date  &  20180805  &  date\_shifted  &  20180805  \\
    2.1... &  Instance  &  (0008,0023)  &  Content Date  &  20180805  &  date\_shifted  &  20180805  \\
    \hline
    \end{tabular}%
  \label{tab:answer_key}%
\end{table*}%

    \subsection{Methods Used for Data Creation}
    
\begin{figure}[htbp]
    \centering
    \includegraphics[width=0.9\linewidth]{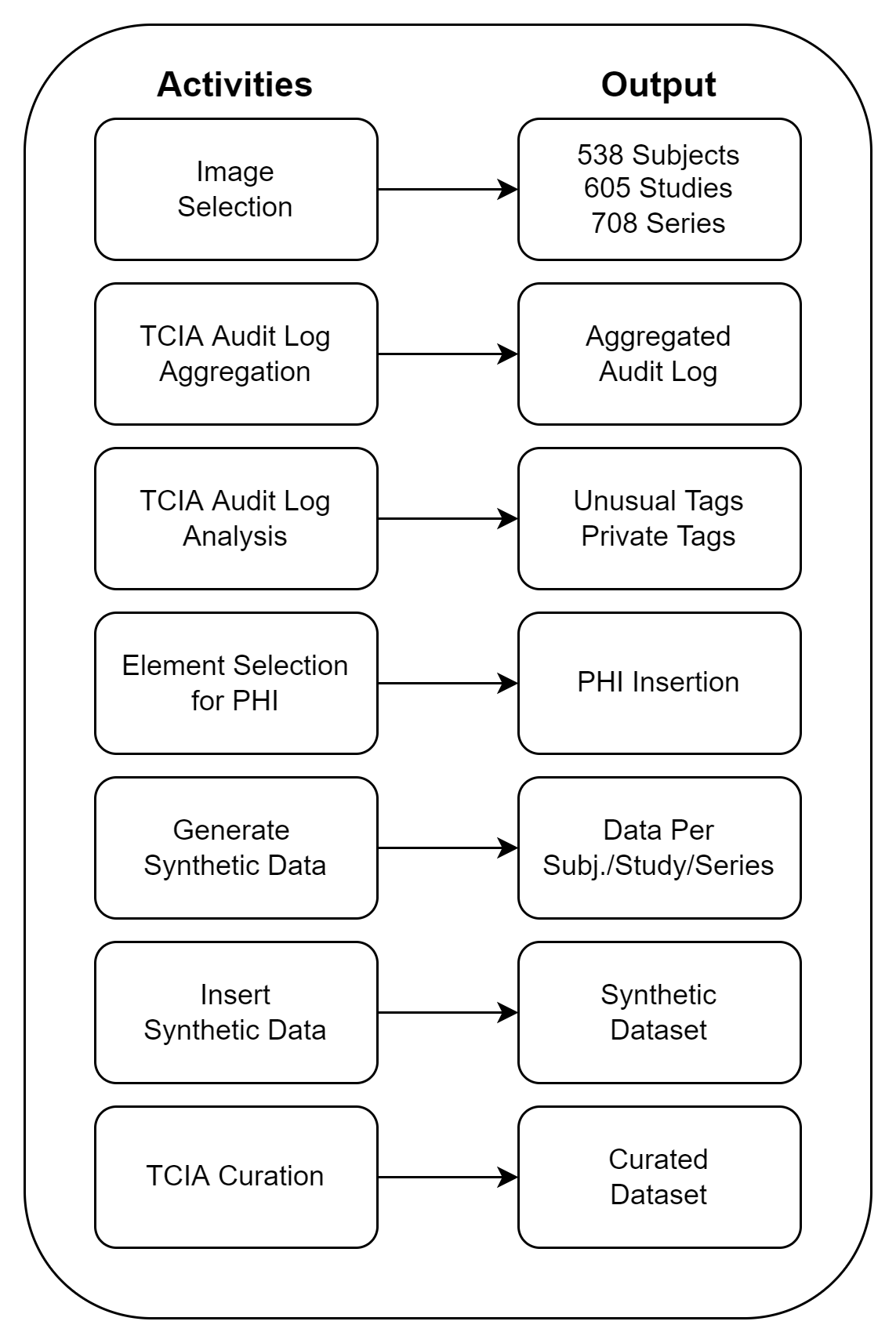}
    \caption{Processing steps for the creation of the datasets.}
    \label{fig:creation_process}
\end{figure}

        The processing steps used to generate our resources are outlined in \textbf{Figure \ref{fig:creation_process}}. To guide the design of our synthetic data, we first analyzed TCIA audit logs. These logs record all curator modifications applied to submitted DICOM files, documenting the provenance of every change, including standard and private data elements, and include the original, replaced, and final values. These curation logs are based predominantly on data submitted using CTP \cite{freymann_image_2012}, which has already performed an initial on-site de-identification covering just over 800 of about 1300 standard DICOM tags, including Patient ID, Patient Name, and Patient Birth Date, which is not logged and original values not accessible to TCIA curators. Submitting sites may have even performed their own in-house de-identification before sending to CTP. Because most common data elements are dealt with before the TCIA curation process, these curation logs are focused more on non-standard DICOM data elements for which curation edits occurred in both typical fields that may contain PHI/PII, like a doctor’s name in a study description, as well as unusual discoveries of PHI/PII. Examples of both such standard and private data elements are listed in our previous work, \cite{rutherford_dicom_2021}.
        
        In addition to the TCIA audit log analysis, we utilized the DICOM Standard and the HIPAA Privacy Rule’s 18 Safe Harbor elements as guidance to identify standard DICOM data elements suitable for synthetic value insertion and address the omission of common fields in the log analysis. This approach ensured comprehensive insertion of synthetic values, including patient names, addresses, and birth dates, into both typical and less common DICOM data elements.

        \subsubsection{Synthetic Data Generation}
            Synthetic PHI/PII was generated using the Python Faker library [\url{https://pypi.org/project/Faker/}], creating pools of synthetic persons, institutions, and addresses. Because some de-identification software tools are able to detect fake addresses as being distinct from actual real addresses, and thence not needing attention, we substituted real addresses from public datasets to increase realism. Records were matched to create pools of synthetic patients and associated imaging studies, including synthetic clinical staff, technicians, and institutions. Synthetic addresses for related entities, such as institutions and physicians, were selected based on geographic proximity at the city, county, and state level to enhance plausibility.

        \subsubsection{Synthetic Data Insertion}
            Synthetic values were inserted into data elements in DICOM files using hand-crafted, rule-based, inheritable templates organized by imaging modality. Each template specified how particular DICOM data elements should be modified, including the re-identification functions (e.g. “generate\_coded\_uid” or “retrieve\_shifted\_date”) to apply for synthetic insertion, relevant parameters such as date shifts assigned to each patient, and the scope of the changes, whether at the study, series, or instance information entity level. Templates were designed to be flexible and editable during synthetic dataset development, allowing for adjustments made directly to the database or import/export spreadsheets. Functions included logic for inserting values under specific conditions, such as only modifying elements that are present or empty, or applying changes to a defined percentage of series, enabling nuanced and variable data generation across diverse imaging contexts.

        \subsubsection{Synthetic Pixel Burn-in}
\begin{figure}[t]
    \centering
    \includegraphics[width=1\linewidth]{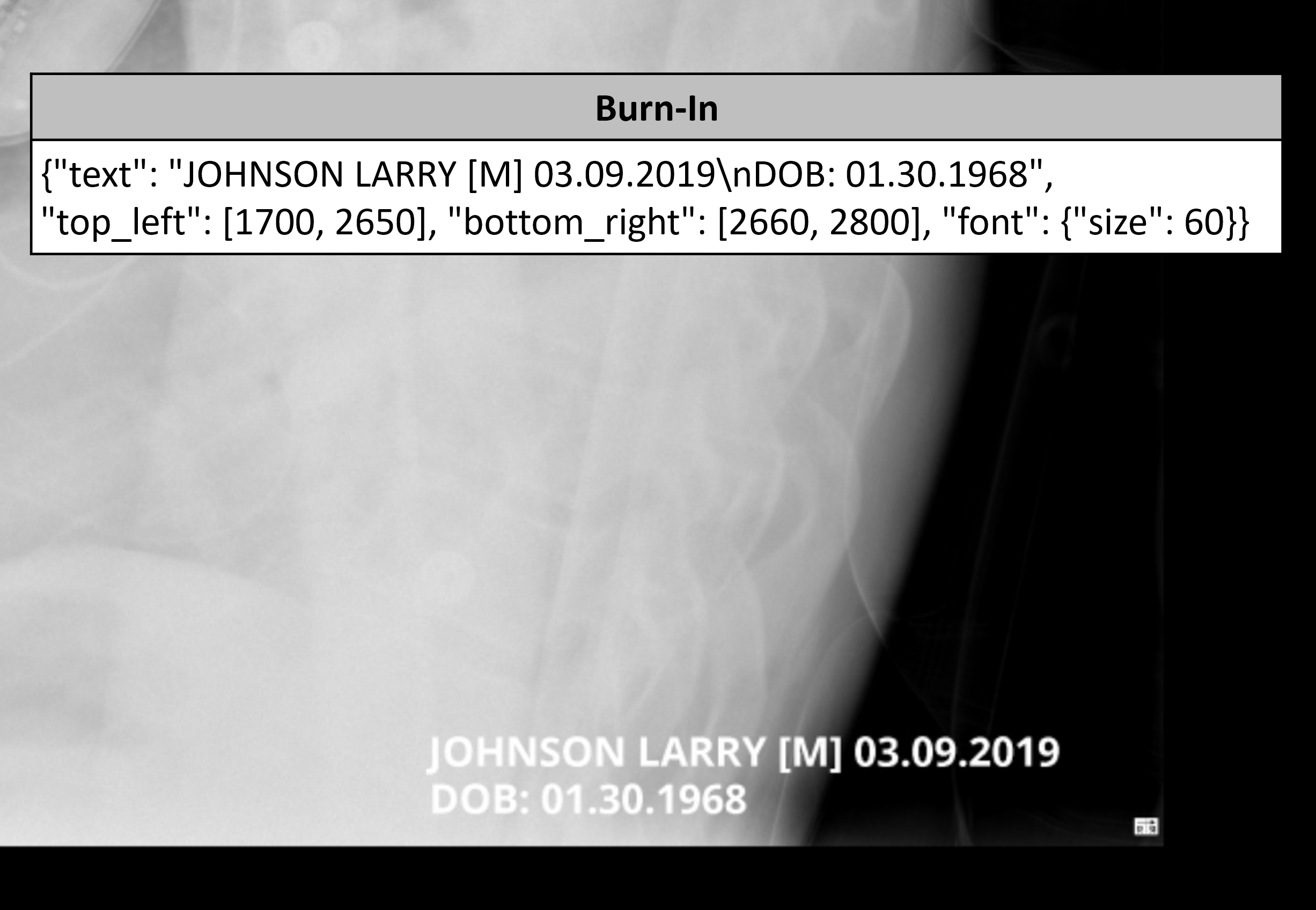}
    \caption{An example of burned-in pixels and the JSON data used to represent it.}
    \label{fig:burned_in_pixels}
\end{figure}

            For a small subset of images, since it rarely occurs in practice for most modalities, PHI/PII was inserted into the pixel data to simulate burned-in text. Images were selected from single-instance series of the CR (Computed Radiography), DX (Digital Radiography), and MG (Mammography) modalities, which historically have a higher likelihood of containing burned-in identifiers, particularly in digitized film or older acquisition workflows. \textbf{Figure \ref{fig:burned_in_pixels}} shows an example image with embedded text and the corresponding JSON object, which defines the text content, bounding box coordinates, and font size.

    \subsection{Software Stack}
\begin{figure}[htbp]
    \centering
    \includegraphics[width=0.4\linewidth]{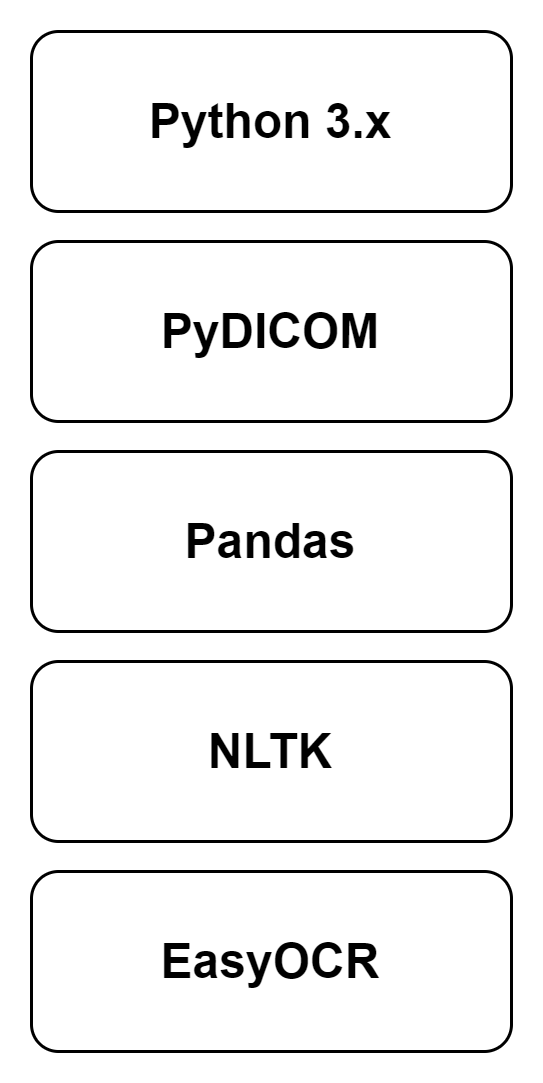}
    \caption{Software Stack.}
    \label{fig:software_stack}
\end{figure}

        The validation software resource was built using Python 3.8 and tested to work with Python 3.9 and 3.10. The resource can be run on Windows, Mac, and Linux and depends on multiple existing well-known Python packages, including NLTK \cite{loper_nltk_2002} for text tokenization, Pandas [\url{https://pandas.pydata.org/}] for in-memory data table management, PyDICOM [\url{https://pydicom.github.io/}] for accessing and editing DICOM images, and EasyOCR [\url{https://github.com/JaidedAI/EasyOCR}] for text detection in the pixel data. A visual representation of this software stack can be found in \textbf{Figure \ref{fig:software_stack}}.

        \subsubsection{MIDI Validation Script}
            The MIDI validation script systematically evaluates whether each de-identified DICOM file from a specific input dataset aligns with the transformations specified in the corresponding answer key. After the synthetic dataset is processed through a de-identification pipeline that is under test, the script traverses every DICOM instance, comparing header elements and pixel data against expected outcomes in the answer key. It checks whether all required actions, such as text removal, date modification, and UID changes, have been correctly applied. For images containing synthetic burned-in text, the script evaluates specific rectangular regions of pixels defined in the answer key using OCR software to confirm that the text was successfully obscured. The entire process produces a comprehensive assessment of de-identification effectiveness and UID mapping consistency across the dataset.        

        \begin{table}[htbp]
  \centering
  \caption{Patient Mapping}
    \begin{tabular}{ll}
    \hline
    \textbf{id\_old} & \textbf{id\_new} \\
    \hline
    31780971 & MIDI\_1\_1\_001 \\
    105230877 & MIDI\_1\_1\_002 \\
    129671781 & MIDI\_1\_1\_003 \\
    205332995 & MIDI\_1\_1\_004 \\
    313822874 & MIDI\_1\_1\_005 \\
    333877207 & MIDI\_1\_1\_006 \\
    356377258 & MIDI\_1\_1\_007 \\
    441491658 & MIDI\_1\_1\_008 \\
    446351014 & MIDI\_1\_1\_009 \\
    527545790 & MIDI\_1\_1\_010 \\
    584796839 & MIDI\_1\_1\_011 \\
    679458347 & MIDI\_1\_1\_012 \\
    \hline
    \end{tabular}%
  \label{tab:patid_mapping}%
\end{table}%

        \subsubsection{Input Files}

            The script expects a configuration file that specifies three main inputs: the de-identified DICOM files, the answer key, and the mapping files. The Patient ID Mapping File (\textbf{Table \ref{tab:patid_mapping}}) links each synthetic patient ID to its corresponding ID in the de-identified dataset, while the UID Mapping File (\textbf{Table \ref{tab:uid_mapping}}) performs the same function for DICOM UIDs. These files ensure traceability between the original and de-identified datasets and answer key to enable consistent evaluation of changes made to the files.
            
            \begin{table}[htbp]
  \centering
  \caption{UID Mapping}
    \begin{tabular}{ll}
    \hline
    \textbf{id\_old} & \textbf{id\_new} \\
    \hline
    3.1.874.1.8955936.308.. & 1.3.6.1.4.1.14519.5.2.1.393.. \\
    3.1.874.1.8955936.119.. & 1.3.6.1.4.1.14519.5.2.1.365.. \\
    3.1.874.1.8955936.192.. & 1.3.6.1.4.1.14519.5.2.1.288.. \\
    3.1.874.1.8955936.327.. & 1.3.6.1.4.1.14519.5.2.1.179.. \\
    3.1.874.1.8955936.257.. & 1.3.6.1.4.1.14519.5.2.1.159.. \\
    3.1.874.1.8955936.663.. & 1.3.6.1.4.1.14519.5.2.1.107.. \\
    3.1.874.1.8955936.191.. & 1.3.6.1.4.1.14519.5.2.1.151.. \\
    3.1.874.1.8955936.395.. & 1.3.6.1.4.1.14519.5.2.1.240.. \\
    3.1.874.1.8955936.325.. & 1.3.6.1.4.1.14519.5.2.1.147.. \\
    3.1.874.1.8955936.107.. & 1.3.6.1.4.1.14519.5.2.1.333.. \\
    3.1.874.1.8955936.865.. & 1.3.6.1.4.1.14519.5.2.1.262.. \\
    3.1.874.1.8955936.249.. & 1.3.6.1.4.1.14519.5.2.1.311.. \\
    \hline
    \end{tabular}%
  \label{tab:uid_mapping}%
\end{table}%
            
            This evaluation process assumes that all Patient ID and UID changes are logged during de-identification, preserving mappings between the original and new identifiers. The tool requires a pseudonymization workflow in which traceability is maintained, rather than a full anonymization process where mappings are permanently removed. However, it can still be applied to test full anonymization pipelines, provided that temporary mappings are available for validation purposes.                       

        \subsubsection{Comparison}
            For each instance, the script identifies relevant actions from the answer key, retrieves the expected value or pattern, and then compares it to the actual value in the de-identified file. For elements like UIDs or Patient IDs, the script ensures changes occur and are applied consistently across all files/instances. Dates are confirmed to be shifted or removed. For text or burned-in pixel actions, it validates whether the string was properly removed, retained, or replaced, and whether pixel regions have been sanitized. Token-based evaluation is used for all textual content, checking each token within a text string individually, potentially resulting in partial per-action scores. Each comparison results in a categorical binary pass or fail (any per-action score below 100\% results in a fail) as well as a continuous per-action score between 0\% and 100\%. Only the binary pass/fail determinations contribute to the final overall dataset scoring and reporting.  

\begin{table}[htbp]
  \centering
  \caption{Discrepancy Report}
    \begin{tabular}{ll}
    \hline    
    \textbf{Header} & \textbf{Description} \\
    \hline    
    check\_passed & Pass / Fail \\
    check\_score & Score between 0 and 100\% \\
    tag\_ds & Tag path in the dataset \\
    tag\_name & Tag name \\
    file\_value & Value in the new file \\
    answer\_value & Value in the original file \\
    action & Action to be taken \\
    action\_text & Text on which the action occurs \\
    category & Answer key category \\
    subcategory & Answer key subcategory \\
    modality & Image modality \\
    class & SOP Class UID \\
    patient & Patient ID \\
    study & Study Instance UID \\
    series & Series Instance UID \\
    instance & SOP Instance UID \\
    file\_name & File name \\
    \hline    
    \end{tabular}%
  \label{tab:discrepancy_report_headers}%
\end{table}%

        \subsubsection{Database}
            All evaluation outcomes are logged at a per-action level in an SQLite database, capturing identifying UIDs, DICOM data element tag path, expected action, de-identified value, pass/fail result, and per-action score. This structure supports post-hoc analysis and enables the consistent generation of summary reports.      

        \begin{table}[htbp]
  \centering
  \caption{Scoring Report}
    \begin{tabular}{ll}
    \hline    
    \textbf{Header} & \textbf{Description} \\
    \hline    
    Category & Category being scored \\
    Fail  & Number of failed actions \\
    Pass  & Number of passed actions \\
    Total & Total number of actions \\
    Score & Final category score \\
    \hline    
    \end{tabular}%
  \label{tab:scoring_report_headers}%
\end{table}%

        \subsubsection{Reporting}
            The script generates multiple pre-defined reports to summarize evaluation results at different levels of detail. Based on a parameter in the configuration file, the reports can be aggregated at either the instance level or series level. For series-level reporting, any discrepancy in any instance within a series for a given action will cause the series to fail a validation check. This aggregation applies to all reports except the discrepancy report.

\begin{table}[htbp]
  \centering
  \caption{Action Report}
    \begin{tabular}{ll}
    \hline    
    \textbf{Header} & \textbf{Description} \\
    \hline    
    Action & Action being scored \\
    Fail  & Number of failed actions \\
    Pass  & Number of passed actions \\
    Total & Total number of actions \\
    \hline    
    \end{tabular}%
  \label{tab:action_report_headers}%
\end{table}%
            
            The Discrepancy Report (\textbf{Table \ref{tab:discrepancy_report_headers}}) lists mismatches between expected and actual values, including the affected SOP Instance UIDs and DICOM data element tag names and values. The Scoring Report (\textbf{Table \ref{tab:scoring_report_headers}}) aggregates the number of passed and failed checks to provide an overall success rate for the dataset being evaluated. The Action Report (\textbf{Table \ref{tab:action_report_headers}}) tracks how often each de-identification action type, such as text\_removed or date\_shifted, was applied and whether it passed or failed, helping to identify error-prone areas. The Category Report (\textbf{Table \ref{tab:category_report_headers}}) maps each action to a higher-level regulatory category. HIPAA, DICOM, or TCIA, specifying the requirements against which each discrepancy is being evaluated.

            \begin{table}[htbp]
  \centering
  \caption{Category Report}
    \begin{tabular}{ll}
    \hline
    \textbf{Header} & \textbf{Description} \\
    \hline    
    Category & Category being scored \\
    Subcategory & Subcategory being scored \\
    Fail  & Number of failed actions \\
    Pass  & Number of passed actions \\
    Total & Total number of actions \\
    \hline    
    \end{tabular}%
  \label{tab:category_report_headers}%
\end{table}%

            Lastly, the Dciodvfy Report (\textbf{Table \ref{tab:dciodvfy_report_headers}}) utilizes the dciodvfy \citep{clunie_dciodvfy_nodate} application to validate the contents of a DICOM file against the DICOM standard rules for encoding and requirements for attributes of information objects. This is in order to confirm that the de-identification process under test makes the files no worse with respect to DICOM conformance. Note that the synthetic dataset is not perfect in this regard, since it was derived from real-world de-identified data.

\begin{table}[htbp]
  \centering
  \caption{Dciodvfy Report}
    \begin{tabular}{ll}
    \hline    
    \textbf{Header} & \textbf{Description} \\
    \hline    
    type  & Warning or Error \\
    tag   & Tag path \\
    message & Descriptive message \\
    modality & Image modality \\
    class & SOP Class UID \\
    patient & Patient ID \\
    study & Study Instance UID \\
    series & Series Instance UID \\
    instance & SOP Instance UID \\
    file\_name & File Name \\
    file\_path & File Path \\
    \hline    
    \end{tabular}%
  \label{tab:dciodvfy_report_headers}%
\end{table}%
            
            Together, these reports provide multiple perspectives on de-identification performance, ranging from individual data element value checks to broader compliance with regulatory and standards-based requirements. The script automates the entire process, from file ingestion and rule application to result logging and report generation, enabling reproducible, high-resolution evaluation of medical image de-identification pipelines.

    \subsection{Flowchart(s) of the Tool's Process Flow}

\begin{figure*}[t]
    \centering
    \includegraphics[width=0.65\linewidth]{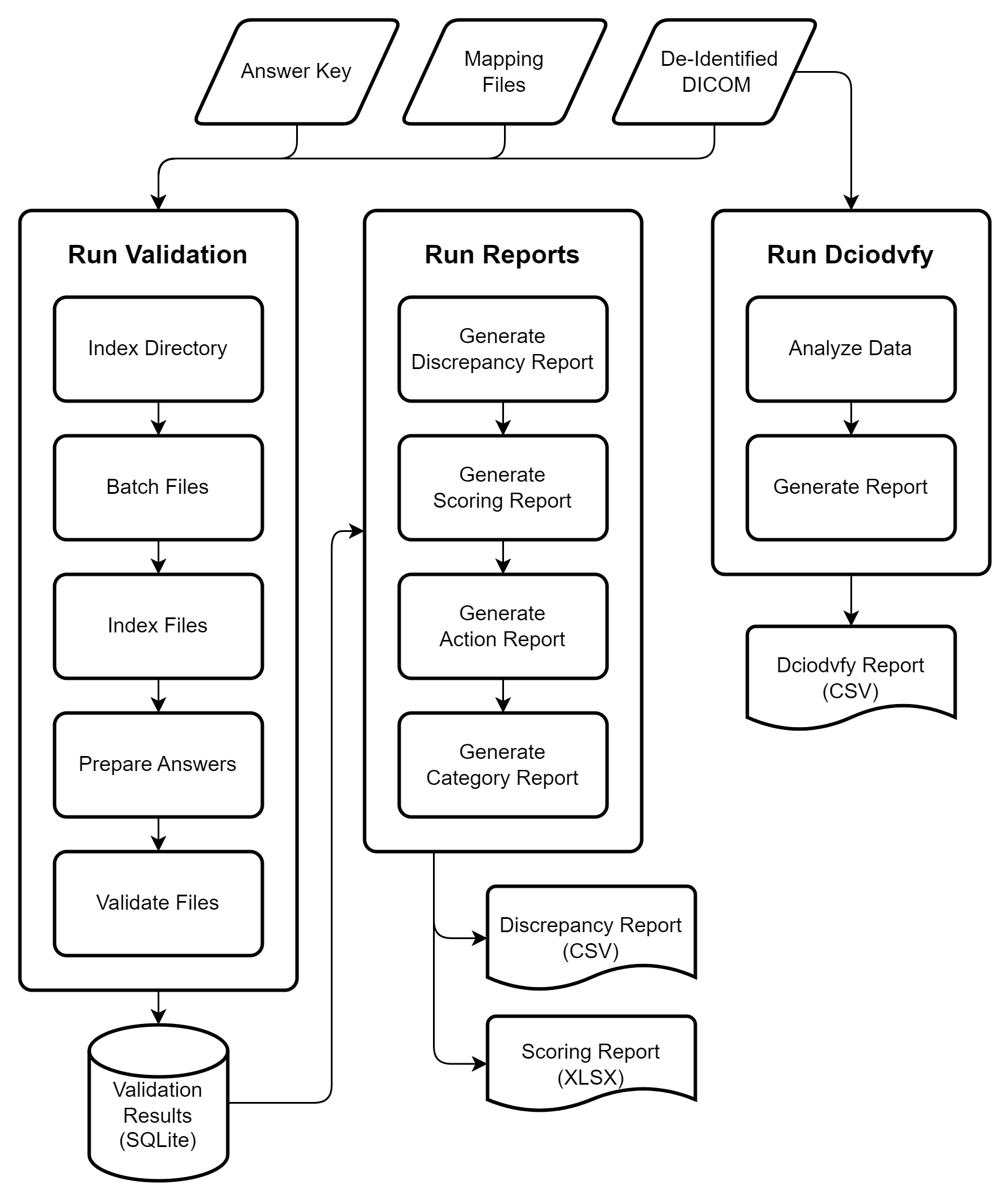}
    \caption{Validation Script Process Flow.}
    \label{fig:software_process}    
\end{figure*}

        \textbf{Figure \ref{fig:software_process}} outlines the process flow of the Validation Script. The Answer Key, Mapping Files, and De-identified DICOM files under test are first ingested by the Run Validation script. This process indexes all the DICOM files in the specified directory to create a list of files. These files are then batched for multi-processing purposes. Files in each batch are parsed to create tables of DICOM data elements and values. For each file being processed, the associated answer key records are prepared and formatted for use in the validation process. The files are then evaluated, and the results are stored in an SQLite database. 

        The Run Reports script is then executed to generate the final reports. This generates in-memory tables for the Discrepancy Report, Scoring Report, Action Report, and Category report. The Discrepancy Report is output as a CSV file, and the Scoring, Action, and Category reports are combined as multiple tabs into an XLSX file called the Scoring Report.
        
        Lastly, the DICOM files are processed through dciodvfy to evaluate DICOM standard conformance. These results are output to a CSV report.
        
    \subsection{Functionalities}
    
        The MIDI evaluation tool provides comprehensive support for evaluating the success of de-identification processes in medical imaging datasets. It evaluates de-identification performance by comparing de-identified DICOM files under test against a known truth answer key, ensuring that all required transformation actions have been correctly applied. The tool performs metadata evaluation with token-by-token scoring, highlighting both successes and deviations at a granular level. It includes detection of residual burned-in text by inspecting pixel regions specified in the answer key, allowing for confirmation of successful obscuration of such text. The tool also checks UID consistency across related DICOM instances to ensure the preservation of referential integrity of relational structures after de-identification. Designed for scalability, the tool supports batch processing of the entire DICOM collection and produces aggregate reports summarizing overall performance at the series or instance level. Output is generated both as a structured SQLite database, enabling downstream programmatic analyses, and as an XLSX report, providing a user-friendly summary suitable for manual review or inclusion in risk analysis documentation. These capabilities offer a standardized, reproducible framework for assessing de-identification workflows against synthetic data infused datasets with known truth, promoting transparency and rigor in data sharing for medical imaging research.         

\section{Validation}
	
    \subsection{Description of Approaches Ensuring the Quality of Data}
        This work involved both the creation of a synthetic medical imaging dataset and the development of an evaluation tool to assess de-identification effectiveness. To ensure the quality, reproducibility, and traceability of the resource, artificial PHI/PII was systematically inserted into both standard and private DICOM data elements, guided by established de-identification practices for medical imaging \cite{moore_-identification_2015,rutherford_dicom_2021}. An answer key was simultaneously generated, automatically recording the precise location and context of each PHI/PII insertion. Manual review of inserted values and answer key entries was performed to confirm completeness and accuracy.
        
        Additional quality assurance involved conducting multiple experimental runs using the Google Cloud Platform (GCP) de-identification product \citep{google_cloud_cloud_nodate}, augmented with project-specific pre- and post-processing scripts, to de-identify the synthetic set as part of the MIDI pipeline project (Opsahl-Ong, et. al. MELBA MIDI special issue). The results of these tests were aggregated and reviewed by an informal group of NCI staff, contractors, and advisors, who identified issues with the GCP de-identification product, the synthetic dataset, and the answer key. These issues were addressed incrementally to refine the dataset and generate the final version of the answer key. 

    \subsection{Experimental Design and Results}
        The MIDI evaluation framework is designed to provide reproducible, standardized assessment of de-identification success in medical imaging datasets. As described in the Methods section, the tool compares de-identified DICOM files against a structured answer key, verifying that prescribed transformations were correctly applied to metadata and pixel data. Each evaluation run produces a SQLite database of comparison outcomes and generates summary reports in both machine-readable and human-readable formats.

        \begin{table}[htbp]
  \centering
  \caption{TCIA Curated Validation Action Report}
    \begin{tabular}{
    >{\raggedright\arraybackslash}p{6em}
    >{\raggedleft\arraybackslash}p{1.5em}
    >{\raggedleft\arraybackslash}p{3.5em}
    >{\raggedleft\arraybackslash}p{3.5em}
    >{\raggedleft\arraybackslash}p{1.5em}
    >{\raggedleft\arraybackslash}p{1.5em}
    }
    \hline
    \textbf{action} & \textbf{Fail} & \textbf{Pass} & \textbf{Total} & \textbf{Pre} & \textbf{Mis} \\
    \hline
    date\_shifted & 0     & 1,502  & 1,502  &       &  \\
    patid\_consistent & 0     & 280   & 280   &       &  \\
    pixels\_hidden & 0     & 31    & 31    &       &  \\
    pixels\_retained & 20    & 229   & 249   &       &  \\
    tag\_retained & 52    & 10,403 & 10,455 & 42    &  \\
    text\_notnull & 73    & 5,867  & 5,940  & 50    &  \\
    text\_removed & 37    & 3,096  & 3,133  &       &  \\
    text\_retained & 461   & 82,814 & 83,275 &       &  \\
    uid\_changed & 0     & 6,402  & 6,402  &       &  \\
    uid\_consistent & 0     & 6,402  & 6,402  &       &  \\
    \hline
          & \textbf{643}   & \textbf{117,026} & \textbf{117,669} & \textbf{92}    & \textbf{0} \\
    \hline          
          &       &       & \textbf{99.45\%} &       &  \\
    \hline          
    \end{tabular}%
  \label{tab:tcia_validation_action}%
\end{table}%
        
        The MIDI evaluation script produces deterministic outputs from standardized inputs, supporting reproducibility across different systems and datasets. Basic validation checks are incorporated to verify the integrity of input files and answer keys prior to execution. All comparisons and scoring operations are deterministic given the same input dataset.

        \begin{table}[htbp]
  \centering
  \caption{TCIA Curated Test Action Report}
    \begin{tabular}{
    >{\raggedright\arraybackslash}p{5.5em}
    >{\raggedleft\arraybackslash}p{2em}
    >{\raggedleft\arraybackslash}p{3.5em}
    >{\raggedleft\arraybackslash}p{3.5em}
    >{\raggedleft\arraybackslash}p{1.5em}
    >{\raggedleft\arraybackslash}p{1.5em}
    }
    \hline
    \textbf{action} & \textbf{Fail} & \textbf{Pass} & \textbf{Total} & \textbf{Pre} & \textbf{Mis} \\
    \hline
    date\_shifted & 0  & 2,306 & 2,306 &    &  \\
    patid\_consistent & 1  & 428 & 429 &    & 1 \\
    pixels\_hidden & 0  & 26 & 26 &    &  \\
    pixels\_retained & 33 & 370 & 403 &    & 1 \\
    tag\_retained & 131 & 15,820 & 15,951 & 63 & 36 \\
    text\_notnull & 128 & 8,910 & 9,038 & 94 & 20 \\
    text\_removed & 246 & 4,491 & 4,737 &    &  \\
    text\_retained & 789 & 126,599 & 127,388 &    & 162 \\
    uid\_changed & 0  & 9,334 & 9,334 &    &  \\
    uid\_consistent & 7  & 9,327 & 9,334 &    &  \\
    \hline
       & \textbf{1,335} & \textbf{177,611} & \textbf{178,946} & \textbf{157} & \textbf{220} \\
    \hline
       &    &    & \textbf{99.25\%} &    &  \\
    \hline
    \end{tabular}%
  \label{tab:tcia_test_action}%
\end{table}%
        
        As an example application, the synthetic dataset was curated by the TCIA curation team and processed through this validation script. In the validation subset, 117,669 checks were evaluated and aggregated at the series level, with 99.45\% passing evaluation (\textbf{Table \ref{tab:tcia_validation_action}}). In the test subset, 178,946 checks were evaluated and aggregated by series, with 99.25\% passing evaluation (\textbf{Table \ref{tab:tcia_test_action}}). Most of the discrepancies were attributed to conservative handling of text\_retained actions, where values were fully removed rather than selectively cleaned, slightly reducing retained data utility but maintaining privacy. Also, some actions related to DICOM attribute type compliance (tag\_retained and text\_notnull) were counted as failures, when the issue occurred in the synthetic dataset and was not addressed. These are listed in the tables as "Pre" for pre-existing. Additional discrepancies arose from the four files removed during the curation process. This is due to the script marking any answer key checks related to data elements in missing files as fails. These are listed in the tables as "Mis" for missing. These overall results demonstrate the tool's effectiveness for producing high-resolution evaluations of de-identification workflows.

\coi{The authors declare no conflicts of interest.}

\acks{This project has been funded in whole or in part with federal funds from the National Cancer Institute, Contract No. 75N91019D00024, Task Orders 75N91020F00003 and 75N91024F00011, and the Google STRIDES contract via Carahsoft, Contract No. OT2OD027060.   The content of this publication does not necessarily reflect the views or policies of the Department of Health and Human Services, nor does mention of trade names, commercial products, or organizations imply endorsement by the U.S. Government.}

\bibliography{midi}

\begin{thebibliography}{22}
\providecommand{\natexlab}[1]{#1}
\providecommand{\url}[1]{\texttt{#1}}
\expandafter\ifx\csname urlstyle\endcsname\relax
  \providecommand{\doi}[1]{doi: #1}\else
  \providecommand{\doi}{doi: \begingroup \urlstyle{rm}\Url}\fi

\bibitem[Bennett et~al.(2018)Bennett, Smith, Jarosz, Nolan, and Bosch]{bennett_reengineering_2018}
William Bennett, Kirk Smith, Quasar Jarosz, Tracy Nolan, and Walter Bosch.
\newblock Reengineering {Workflow} for {Curation} of {DICOM} {Datasets}.
\newblock \emph{Journal of Digital Imaging}, 31\penalty0 (6):\penalty0 783--791, December 2018.
\newblock ISSN 0897-1889, 1618-727X.
\newblock \doi{10.1007/s10278-018-0097-4}.
\newblock URL \url{http://link.springer.com/10.1007/s10278-018-0097-4}.

\bibitem[Clark et~al.(2013)Clark, Vendt, Smith, Freymann, Kirby, Koppel, Moore, Phillips, Maffitt, Pringle, Tarbox, and Prior]{clark_cancer_2013}
Kenneth Clark, Bruce Vendt, Kirk Smith, John Freymann, Justin Kirby, Paul Koppel, Stephen Moore, Stanley Phillips, David Maffitt, Michael Pringle, Lawrence Tarbox, and Fred Prior.
\newblock The {Cancer} {Imaging} {Archive} ({TCIA}): {Maintaining} and {Operating} a {Public} {Information} {Repository}.
\newblock \emph{Journal of Digital Imaging}, 26\penalty0 (6):\penalty0 1045--1057, December 2013.
\newblock ISSN 0897-1889, 1618-727X.
\newblock \doi{10.1007/s10278-013-9622-7}.
\newblock URL \url{http://link.springer.com/10.1007/s10278-013-9622-7}.

\bibitem[Clunie()]{clunie_dciodvfy_nodate}
David~A. Clunie.
\newblock dciodvfy.
\newblock URL \url{https://www.dclunie.com/dicom3tools/dciodvfy.html}.

\bibitem[Clunie et~al.(2025)Clunie, Flanders, Taylor, Erickson, Bialecki, Brundage, Gutman, Prior, Seibert, Perry, Gichoya, Kirby, Andriole, Geneslaw, Moore, Fitzgerald, Tellis, Xiao, Farahani, Luo, Rosenthal, Kandarpa, Rosen, Goetz, Babcock, Xu, and Hsiao]{clunie_report_2025}
David~A. Clunie, Adam Flanders, Adam Taylor, Brad Erickson, Brian Bialecki, David Brundage, David Gutman, Fred Prior, J.~Anthony Seibert, John Perry, Judy~Wawira Gichoya, Justin Kirby, Katherine Andriole, Luke Geneslaw, Steve Moore, T.~J. Fitzgerald, Wyatt Tellis, Ying Xiao, Keyvan Farahani, James Luo, Alex Rosenthal, Kris Kandarpa, Rebecca Rosen, Kerry Goetz, Debra Babcock, Ben Xu, and John Hsiao.
\newblock Report of the {Medical} {Image} {De}-{Identification} ({MIDI}) {Task} {Group} - {Best} {Practices} and {Recommendations}.
\newblock \emph{ArXiv}, page arXiv:2303.10473v3, March 2025.
\newblock ISSN 2331-8422.
\newblock URL \url{https://doi.org/10.48550/arXiv.2303.10473}.

\bibitem[Dankar and Ibrahim(2021)]{dankar_fake_2021}
Fida~K. Dankar and Mahmoud Ibrahim.
\newblock Fake {It} {Till} {You} {Make} {It}: {Guidelines} for {Effective} {Synthetic} {Data} {Generation}.
\newblock \emph{Applied Sciences}, 11\penalty0 (5):\penalty0 2158, February 2021.
\newblock ISSN 2076-3417.
\newblock \doi{10.3390/app11052158}.
\newblock URL \url{https://www.mdpi.com/2076-3417/11/5/2158}.

\bibitem[De~Kok et~al.(2023)De~Kok, De~La~Hoz, De~Jong, Brokke, Elbers, Thoral, Castillejo, Trenor, Castellano, Bronchalo, Merz, Faltys, {Collaborator group}, Casares, Jiménez, Requejo, Gutiérrez, Curto, Rätsch, Peppink, Driessen, Sijbrands, Kompanje, Girbes, Barberan, Varona, Villares, Van Der~Horst, Xu, Celi, Van~Bussel, and Borrat]{de_kok_guide_2023}
Jip W. T.~M. De~Kok, Miguel Á.~Armengol De~La~Hoz, Ymke De~Jong, Véronique Brokke, Paul W.~G. Elbers, Patrick Thoral, Alejandro Castillejo, Tomás Trenor, Jose~M. Castellano, Alberto~E. Bronchalo, Tobias~M. Merz, Martin Faltys, {Collaborator group}, Cristina Casares, Araceli Jiménez, Jaime Requejo, Sonia Gutiérrez, David Curto, Gunnar Rätsch, Jan~M. Peppink, Ronald~H. Driessen, Eric J.~G. Sijbrands, Erwin J.~O. Kompanje, Armand R.~J. Girbes, Jose Barberan, Jose~Felipe Varona, Paula Villares, Iwan C.~C. Van Der~Horst, Minnan Xu, Leo~Anthony Celi, Bas C.~T. Van~Bussel, and Xavier Borrat.
\newblock A guide to sharing open healthcare data under the {General} {Data} {Protection} {Regulation}.
\newblock \emph{Scientific Data}, 10\penalty0 (1):\penalty0 404, June 2023.
\newblock ISSN 2052-4463.
\newblock \doi{10.1038/s41597-023-02256-2}.
\newblock URL \url{https://www.nature.com/articles/s41597-023-02256-2}.

\bibitem[Diaz et~al.(2021)Diaz, Kushibar, Osuala, Linardos, Garrucho, Igual, Radeva, Prior, Gkontra, and Lekadir]{diaz_data_2021}
Oliver Diaz, Kaisar Kushibar, Richard Osuala, Akis Linardos, Lidia Garrucho, Laura Igual, Petia Radeva, Fred Prior, Polyxeni Gkontra, and Karim Lekadir.
\newblock Data preparation for artificial intelligence in medical imaging: {A} comprehensive guide to open-access platforms and tools.
\newblock \emph{Physica Medica}, 83:\penalty0 25--37, March 2021.
\newblock ISSN 11201797.
\newblock \doi{10.1016/j.ejmp.2021.02.007}.
\newblock URL \url{https://linkinghub.elsevier.com/retrieve/pii/S1120179721000958}.

\bibitem[DuMont~Schütte et~al.(2021)DuMont~Schütte, Hetzel, Gatidis, Hepp, Dietz, Bauer, and Schwab]{dumont_schutte_overcoming_2021}
August DuMont~Schütte, Jürgen Hetzel, Sergios Gatidis, Tobias Hepp, Benedikt Dietz, Stefan Bauer, and Patrick Schwab.
\newblock Overcoming barriers to data sharing with medical image generation: a comprehensive evaluation.
\newblock \emph{npj Digital Medicine}, 4\penalty0 (1):\penalty0 141, December 2021.
\newblock ISSN 2398-6352.
\newblock \doi{10.1038/s41746-021-00507-3}.
\newblock URL \url{https://www.nature.com/articles/s41746-021-00507-3}.

\bibitem[Freymann et~al.(2012)Freymann, Kirby, Perry, Clunie, and Jaffe]{freymann_image_2012}
John~B. Freymann, Justin~S. Kirby, John~H. Perry, David~A. Clunie, and C.~Carl Jaffe.
\newblock Image {Data} {Sharing} for {Biomedical} {Research}—{Meeting} {HIPAA} {Requirements} for {De}-identification.
\newblock \emph{Journal of Digital Imaging}, 25\penalty0 (1):\penalty0 14--24, February 2012.
\newblock ISSN 0897-1889, 1618-727X.
\newblock \doi{10.1007/s10278-011-9422-x}.
\newblock URL \url{http://link.springer.com/10.1007/s10278-011-9422-x}.

\bibitem[Gonzales et~al.(2023)Gonzales, Guruswamy, and Smith]{gonzales_synthetic_2023}
Aldren Gonzales, Guruprabha Guruswamy, and Scott~R. Smith.
\newblock Synthetic data in health care: {A} narrative review.
\newblock \emph{PLOS Digital Health}, 2\penalty0 (1):\penalty0 e0000082, January 2023.
\newblock ISSN 2767-3170.
\newblock \doi{10.1371/journal.pdig.0000082}.
\newblock URL \url{https://dx.plos.org/10.1371/journal.pdig.0000082}.

\bibitem[{Google Cloud}()]{google_cloud_cloud_nodate}
{Google Cloud}.
\newblock Cloud {Healthcare} {API} documentation.
\newblock URL \url{https://cloud.google.com/healthcare-api/docs}.

\bibitem[Kondylakis et~al.(2024)Kondylakis, Catalan, Alabart, Barelle, Bizopoulos, Bobowicz, Bona, Fotiadis, Garcia, Gomez, Jimenez-Pastor, Karatzanis, Lekadir, Kogut-Czarkowska, Lalas, Marias, Marti-Bonmati, Munuera, Nikiforaki, Pelissier, Prior, Rutherford, Saint-Aubert, Sakellariou, Seymour, Trouillard, Votis, and Tsiknakis]{kondylakis_documenting_2024}
Haridimos Kondylakis, Rocio Catalan, Sara~Martinez Alabart, Caroline Barelle, Paschalis Bizopoulos, Maciej Bobowicz, Jonathan Bona, Dimitrios~I. Fotiadis, Teresa Garcia, Ignacio Gomez, Ana Jimenez-Pastor, Giannis Karatzanis, Karim Lekadir, Magdalena Kogut-Czarkowska, Antonios Lalas, Kostas Marias, Luis Marti-Bonmati, Jose Munuera, Katerina Nikiforaki, Manon Pelissier, Fred Prior, Michael Rutherford, Laure Saint-Aubert, Zisis Sakellariou, Karine Seymour, Thomas Trouillard, Konstantinos Votis, and Manolis Tsiknakis.
\newblock Documenting the de-identification process of clinical and imaging data for {AI} for health imaging projects.
\newblock \emph{Insights into Imaging}, 15\penalty0 (1):\penalty0 130, May 2024.
\newblock ISSN 1869-4101.
\newblock \doi{10.1186/s13244-024-01711-x}.
\newblock URL \url{https://insightsimaging.springeropen.com/articles/10.1186/s13244-024-01711-x}.

\bibitem[Kushida et~al.(2012)Kushida, Nichols, Jadrnicek, Miller, Walsh, and Griffin]{kushida_strategies_2012}
Clete~A. Kushida, Deborah~A. Nichols, Rik Jadrnicek, Ric Miller, James~K. Walsh, and Kara Griffin.
\newblock Strategies for de-identification and anonymization of electronic health record data for use in multicenter research studies.
\newblock \emph{Medical care}, 50\penalty0 (Suppl):\penalty0 S82--101, July 2012.
\newblock ISSN 0025-7079.
\newblock \doi{10.1097/MLR.0b013e3182585355}.
\newblock URL \url{https://www.ncbi.nlm.nih.gov/pmc/articles/PMC6502465/}.

\bibitem[Langlois et~al.(2024)Langlois, Szelagowski, Vanderdonckt, and Jodogne]{langlois_open_2024}
Quentin Langlois, Nicolas Szelagowski, Jean Vanderdonckt, and Sébastien Jodogne.
\newblock Open {Platform} for the {De}-identification of {Burned}-in {Texts} in {Medical} {Images} using {Deep} {Learning}:.
\newblock In \emph{Proceedings of the 17th {International} {Joint} {Conference} on {Biomedical} {Engineering} {Systems} and {Technologies}}, pages 297--304, Rome, Italy, 2024. SCITEPRESS - Science and Technology Publications.
\newblock ISBN 978-989-758-688-0.
\newblock \doi{10.5220/0012430300003657}.
\newblock URL \url{https://www.scitepress.org/DigitalLibrary/Link.aspx?doi=10.5220/0012430300003657}.

\bibitem[Loper and Bird(2002)]{loper_nltk_2002}
Edward Loper and Steven Bird.
\newblock {NLTK}: {The} {Natural} {Language} {Toolkit}, 2002.
\newblock URL \url{https://arxiv.org/abs/cs/0205028}.
\newblock Version Number: 1.

\bibitem[Moore et~al.(2015)Moore, Maffitt, Smith, Kirby, Clark, Freymann, Vendt, Tarbox, and Prior]{moore_-identification_2015}
Stephen~M. Moore, David~R. Maffitt, Kirk~E. Smith, Justin~S. Kirby, Kenneth~W. Clark, John~B. Freymann, Bruce~A. Vendt, Lawrence~R. Tarbox, and Fred~W. Prior.
\newblock De-identification of {Medical} {Images} with {Retention} of {Scientific} {Research} {Value}.
\newblock \emph{RadioGraphics}, 35\penalty0 (3):\penalty0 727--735, May 2015.
\newblock ISSN 0271-5333, 1527-1323.
\newblock \doi{10.1148/rg.2015140244}.
\newblock URL \url{http://pubs.rsna.org/doi/10.1148/rg.2015140244}.

\bibitem[{NEMA}({\natexlab{a}})]{nema_digital_nodate}
{NEMA}.
\newblock Digital {Imaging} and {Communications} in {Medicine} ({DICOM}) {Standard}, {\natexlab{a}}.
\newblock URL \url{http://www.dicomstandard.org/}.

\bibitem[{NEMA}({\natexlab{b}})]{nema_digital_nodate-1}
{NEMA}.
\newblock Digital {Imaging} and {Communications} in {Medicine} ({DICOM}) {Standard} {PS3}.15: {Security} and {System} {Management} {Profiles}, {\natexlab{b}}.
\newblock URL \url{http://dicom.nema.org/}.

\bibitem[Pezoulas et~al.(2024)Pezoulas, Zaridis, Mylona, Androutsos, Apostolidis, Tachos, and Fotiadis]{pezoulas_synthetic_2024}
Vasileios~C. Pezoulas, Dimitrios~I. Zaridis, Eugenia Mylona, Christos Androutsos, Kosmas Apostolidis, Nikolaos~S. Tachos, and Dimitrios~I. Fotiadis.
\newblock Synthetic data generation methods in healthcare: {A} review on open-source tools and methods.
\newblock \emph{Computational and Structural Biotechnology Journal}, 23:\penalty0 2892--2910, December 2024.
\newblock ISSN 2001-0370.
\newblock \doi{10.1016/j.csbj.2024.07.005}.
\newblock URL \url{https://www.sciencedirect.com/science/article/pii/S2001037024002393}.

\bibitem[Phillips and Knoppers(2016)]{phillips_discombobulation_2016}
Mark Phillips and Bartha~M Knoppers.
\newblock The discombobulation of de-identification.
\newblock \emph{Nature Biotechnology}, 34\penalty0 (11):\penalty0 1102--1103, November 2016.
\newblock ISSN 1087-0156, 1546-1696.
\newblock \doi{10.1038/nbt.3696}.
\newblock URL \url{http://www.nature.com/articles/nbt.3696}.

\bibitem[Rutherford et~al.(2021)Rutherford, Mun, Levine, Bennett, Smith, Farmer, Jarosz, Wagner, Freyman, Blake, Tarbox, Farahani, and Prior]{rutherford_dicom_2021}
Michael Rutherford, Seong~K. Mun, Betty Levine, William Bennett, Kirk Smith, Phil Farmer, Quasar Jarosz, Ulrike Wagner, John Freyman, Geri Blake, Lawrence Tarbox, Keyvan Farahani, and Fred Prior.
\newblock A {DICOM} dataset for evaluation of medical image de-identification.
\newblock \emph{Scientific Data}, 8\penalty0 (1):\penalty0 183, December 2021.
\newblock ISSN 2052-4463.
\newblock \doi{10.1038/s41597-021-00967-y}.
\newblock URL \url{http://www.nature.com/articles/s41597-021-00967-y}.

\bibitem[{U.S. Dept. of Health and Human Services}(2012)]{us_dept_of_health_and_human_services_guidance_2012}
{U.S. Dept. of Health and Human Services}.
\newblock Guidance {Regarding} {Methods} for {De}-identification of {Protected} {Health} {Information} in {Accordance} with the {Health} {Insurance} {Portability} and {Accountability} {Act} ({HIPAA}) {Privacy} {Rule}, 2012.
\newblock URL \url{https://www.hhs.gov/hipaa/for-professionals/privacy/special-topics/de-identification/index.html}.

\end{thebibliography}




\end{document}